\documentclass{article}
\usepackage{ijcai23}

\usepackage{times}
\usepackage{soul}
\usepackage{url}
\usepackage[hidelinks]{hyperref}
\usepackage[utf8]{inputenc}
\usepackage[small]{caption}
\usepackage{graphicx}
\usepackage{amsmath}
\usepackage{amsthm}
\usepackage{booktabs}
\usepackage{algorithm}
\usepackage{algorithmic}
\usepackage[switch]{lineno}

\usepackage{subcaption}
\usepackage{xcolor}
\usepackage{bbding}
\usepackage{xspace}
\usepackage{placeins}

\newcommand{\Fig}{Fig.\xspace}
\newcommand{\Tab}{Tab.\xspace}

\newcommand{\eg}{e.g.}
\newcommand{\ie}{i.e.\xspace}

\newcommand{\tracker}{DA-Tracker\xspace}

\title{Tracking Different Ant Species: An Unsupervised Domain Adaptation Framework and a Dataset for Multi-object Tracking}

\author{
Chamath Abeysinghe$^1$
\and
Chris Reid$^2$\and
Hamid Rezatofighi$^1$\And
Bernd Meyer$^1$
\affiliations
$^1$Dept. of Data Science and Artificial Intelligence, Monash University\\
$^2$School of Natural Sciences, Macquarie University\\
\emails
\{chamath.abeysinghe, hamid.rezatofighi,  bernd.meyer\}@monash.edu,
chris.reid@mq.edu.au
}

\begin{document}

\maketitle

\begin{abstract}
    Tracking individuals is a vital part of many experiments conducted to understand collective  behaviour. Ants are the paradigmatic model system for such experiments but their lack of individually distinguishing visual features and their high colony densities make it extremely difficult to perform reliable  tracking automatically. Additionally, the wide diversity of their species'  appearances makes a generalized approach even harder. In this paper, we propose a data-driven multi-object tracker that, for the first time, employs domain adaptation to achieve the required generalisation. This approach is built upon a joint-detection-and-tracking framework that is extended by a set of domain discriminator modules integrating an adversarial training strategy in addition to the tracking loss. In addition to this novel domain-adaptive tracking framework, we present a new dataset and a benchmark for the ant tracking problem. The dataset contains 57 video sequences with full trajectory annotation, including 30k frames captured from two different ant species moving on  different background patterns. It comprises 33 and 24 sequences for source and target domains, respectively. We compare our proposed framework against other domain-adaptive and non-domain-adaptive multi-object tracking baselines using this dataset and show that incorporating domain adaptation at multiple levels of the tracking pipeline yields significant improvements. The code and the dataset are available at \textcolor{magenta}{\emph{https://github.com/chamathabeysinghe/da-tracker}}.
\end{abstract}

\section{Introduction}

Many biologists and ecologists are interested in understanding social insect behaviours, in particular, that of ants to gain insights into how social systems collaborate in nature. This important research is unfortunately slowed down by the difficulty of automatically tracking ants of many different species in very diverse experimental environments.

Compared to other multi-object tracking (MOT) problems in computer vision, ant tracking imposes unique challenges; individual ants are visually extremely similar and move in highly crowded environments with  complex interactions. This leads to a significant level of occlusions, overlapping movements \emph{etc}. 
Furthermore, ants comprise a great variety of species with broadly differing appearances between species. The main problem for data-driven approaches arising from this diversity is the transferability between species. State-of-the-art
data-driven tracking methods such as~\cite{trackformer,transtrack} trained on a dataset of a specific ant species and environment do not perform well on   datasets with different ant species and environments, as  shown in the experiment section of this paper. Consequently, the most popular approaches used for ant tracking rely on model-based detection and tracking techniques~\cite{idtrackerv1,naiser2018tracking}. To the best of our knowledge, no species-diverse large-scale dataset exists that can be used for training and comparing state-of-the-art data-driven detection and tracking approaches, \eg,~\cite{fasterrcnn,trackformer,transtrack}.


\begin{figure}[tb!]
\begin{center}
    
\begin{subfigure}[b]{0.47\textwidth}
    \centering
         \includegraphics[width=0.8\textwidth]{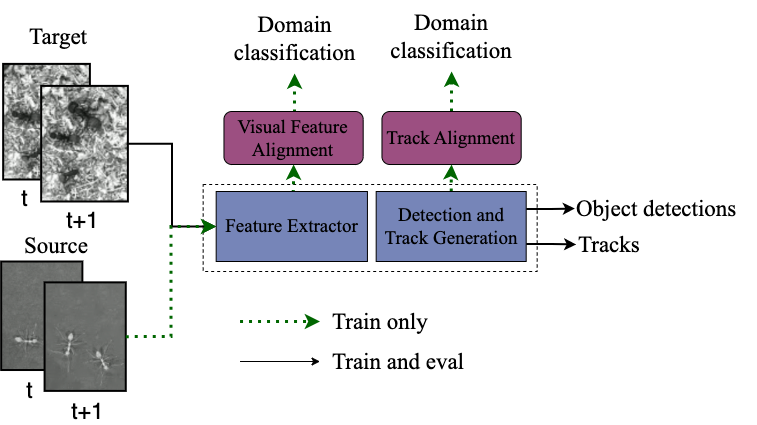}
         \caption{}
         \label{fig:simiplifiedarchitecture}
\end{subfigure}

\begin{subfigure}[b]{0.45\textwidth}
    \centering
         \includegraphics[width=\textwidth]{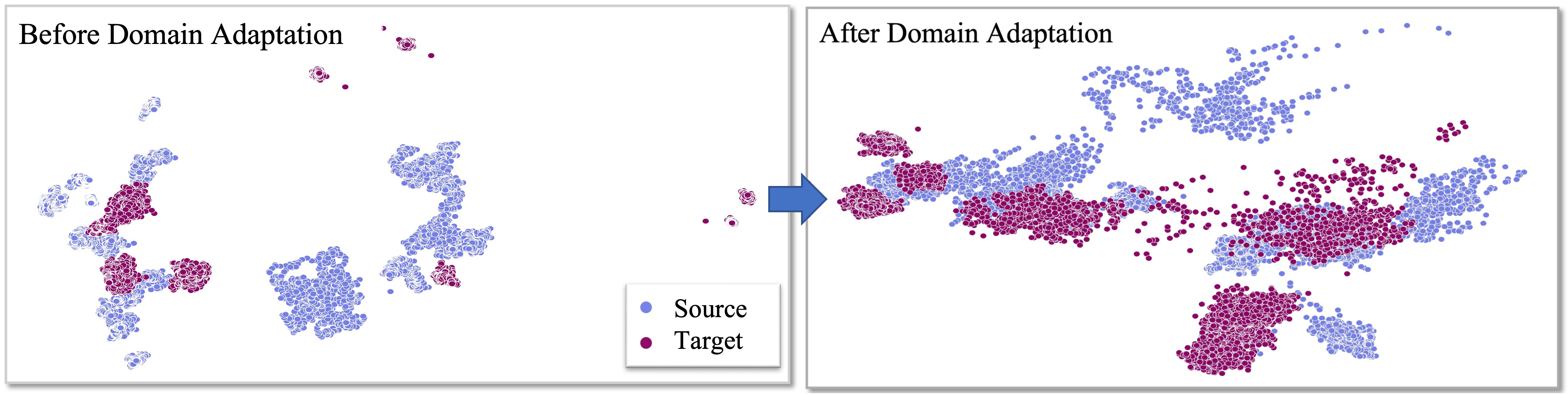}
         \caption{}
         \label{fig:beforeafterdomainadapation}
\end{subfigure}

\end{center}
\vspace{-1em}
\caption{ (a) A schematic of our proposed unsupervised domain adaptation framework for multi-object tracking, (b) A visualization, representing the source and target feature distributions before and after applying our domain adaptation strategy. }

\label{fig:intro_overview}
\end{figure}

In this paper, we propose a data-driven domain adaptive multi-object tracking framework, named \tracker, which is capable of transferring knowledge from training with one ant species (source) to another ant species in a different environment (target). To the best of our knowledge, our proposed approach is the first MOT framework to tackle this unsupervised domain adaptation problem in an end-to-end trainable manner. \Fig~\ref{fig:intro_overview} shows a high-level design of the proposed framework and its performance in unifying the source and target feature representation.   

Our proposed end-to-end multi-object tracking approach builds upon~\cite{trackformer} and extends this to perform domain adaptation. Our domain adaptation modules are based on adversarial training strategies that generate similarity between the encoded input and decoded output representations from source and target data by enforcing highly overlapping feature distributions. 
To this end, domain adaptation is applied to image level features and track level information separately. To train \tracker, we use two different categories of losses: 1) supervised losses, used for source domain data only, and 2) discriminator losses, used for both source and target domain data. The multi-object tracking module learns to localize objects and generate tracks using the supervised losses from the annotated outputs available in the source data. The three discriminators, connected to the intermediate encoding and decoding layers in the tracking pipeline, ensure that the tracking module generates highly similar  feature distributions for source and target domains data across the input and output representations.


As a second contribution, we introduce and make publicly available a new large-scale dataset for ant tracking with unsupervised domain adaptation in realistic experimental setups. We use this to evaluate our framework in this paper. Our dataset includes 57 video sequences (more than $30K$ image frames) captured in a laboratory environment from  two ant species {\it Oecophylla smaragdina} (weaver ants)  and {\it Camponotus aeneopilosus} (carpenter ants) with very different appearance and  behaviour  moving on different background structures (ranging from plain to grassy backgrounds) under various ant population density, lighting and the camera zoom conditions (\Fig~\ref{fig:data-examples}). The dataset is provided with about $700K$ and $2K$ high-quality bounding boxes and tracks, respectively. To establish a standard benchmark, we divide both the target and source domain data into three splits: training, validation and test. The training split comprises about $50\%$ of data; test and validation about $25\%$ each. To evaluate our method, we adopt standard multi-object tracking metrics such as MOTClear~\cite{clearmotmetric}, IDF~\cite{idfmetric} and HOTA~\cite{hotametric}.




In summary, our main contributions are to: (1) propose an unsupervised domain adaptation method for multi-object tracking in an end-to-end trainable framework, (2) introduce a large-scale ant dataset and benchmark for unsupervised domain adaptation in multi-object tracking, (3) comprehensively evaluate and compare our framework against the state-of-the-art detection based and data-driven MOT approaches.

\section{Related work}

\textbf{Multi-object tracking in computer vision:} 
In multi-object tracking problems, data-driven approaches have shown state-of-the-art performance on many publicly available MOT datasets such as~\cite{motchallenge16,kittidataset}. 
A main paradigm in multi-object tracking is tracking-by-detection where the problem is dived into a two-step process: (1) object detection and (2) track generation. In the first step, a deep learning-based object detector, \eg,~\cite{fasterrcnn,maskrcnn,detr}, is used to localize all the objects of interest in all the frame sequences. Next, a tracking technique is applied to generate tracks, \eg, a filtering-based framework~\cite{sortalgorithm,deepsort,rezatofighi2015joint}, an optimisation-based technique~\cite{optimizetrack1,optimizetrack4ijcai} 
or another approach~\cite{othertrack1,othertrack2} using a similarity/distance measure between detection and hypothetical tracks, that is generally based on appearance information~\cite{deepsort,appearanceinfo3ijcai}, motion information~\cite{motioninfo2,motioninfo3} or both~\cite{bothinfo1,bothinfo2}. 


A different multi-object tracking paradigm that has recently become popular is joint-detection-and-tracking~\cite{trackformer,transtrack,tracktor,jointtrack1}. 
Joint-detection-and-tracking performs object detection and tracking simultaneously in a single step. Integrating everything into a single step allows us to efficiently exchange information between object detection and tracking. Our proposed model is built on top of this category of MOT approaches. In this paper, we use Trackformer~\cite{trackformer} as an end-to-end trainable MOT framework as our baseline. Trackformer~\cite{trackformer} is a joint-detection-and-tracking method inspired by the transformer object detector, DETR~\cite{detr}. Trackformer has two object detectors working on consecutive frames. It learns to initialize new tracks (\ie new detected objects), and terminate the exiting tracks or associate them to the next frame by predicting their next state  (\ie bounding box and confidence scores) in an auto-regressive scheme between two consecutive frames.

\textbf{Ant tracking:} 
Most existing ant tracking frameworks follow model-based approaches~\cite{idtrackerv1,naiser2018tracking}.  
A popular monitoring tool used for ant tracking is iDTracker. As originally described in~\cite{idtrackerv1}, this tool generates object detections using colour intensity thresholds and assigns a unique ID using a classification network. The original approach is model-based and does not need training data. It can efficiently work under simple tracking scenarios. However, it fails to track reliably in highly crowded environments. We note that a more recent extension, termed iDTracker.ai, uses deep learning (CNNs) for individual identification~\cite{idtrackerv2}. This approach relies on individuals being sufficiently distinguishable and no results for ants have been published in~\cite{idtrackerv2}. Another tracking-by-detection approach using deep learning  for object detection (Mask R-CNN~\cite{maskrcnn}) is described in~\cite{naturemethodsanttracking}. This approach uses Earth Mover’s Distance (EMD)~\cite{emd} to link detections into tracks. The performance of these tracking-by-detection approaches heavily relies on the supervised object detectors which need to be re-trained for every experimental setting.

\textbf{Unsupervised domain adaptation:} 
Unsupervised domain adaptation is a process of generalizing a model to work on other input data than the labeled training data. There are many approaches to unsupervised domain adaptation, \eg\ adversarial-based methods~\cite{strongweakfeaturealignment,da_frcnn}, discrepancy-based methods~\cite{discrepencybased1,discrepencybased3}, and reconstruction-based methods~\cite{reconstructionbased1,reconstructionbased3}.
Domain adaptation is a very well-studied problem in a few machine learning/computer vision tasks such as image classification ~\cite{grl,da_image_classification3ijcai} and object detection ~\cite{strongweakfeaturealignment,da_frcnn,da_detr1,da_transformersijcai}. However, it has been barely extended to higher-level tasks, such as multi-object tracking, due to the complexity of such problem.

Many domain adaptation methods~\cite{da_frcnn,strongweakfeaturealignment,da_detr1,da_transformersijcai} in object detection are adversarial-based methods that train image domain classifiers alongside the object detector. In adversarial-based domain adaptation, model generalization is achieved by a discriminator forcing the feature extractor to translate feature representations for different domain data into one common  distribution.
\cite{da_frcnn} have proposed a Faster R-CNN  with an adversarial domain adaptive approach. In this method, a Faster R-CNN  connects to domain classification layers at two levels: the image level and the instance level, and translates features into a common distribution. \cite{strongweakfeaturealignment} propose improvements to this DA-Faster R-CNN method by introducing a new loss function that estimates a cost based on the classification's difficulty.

Only a few studies of domain adaptation in the multi-object tracking exist~\cite{da_tracker1}. One way to improve tracking's accuracy for an unseen domain is to use a domain-adaptive object detector followed by a model-based tracking approach. The baseline models to which we compare in this study work in this way. In contrast to this and previous work, we apply domain adaptation techniques on both object detection and track generation in a data-driven multi-object tracker.

\textbf{Dataset and benchmark:} Several large-scale standard datasets and benchmarks for multi-object tracking problems exist, \eg\  for pedestrian/human tracking~\cite{motchallenge16,pdestre,dancetrack}, vehicle tracking~\cite{kittidataset,uavdt}, and animal tracking~\cite{argos}. Having these public datasets and benchmarks helps to develop state-of-the-art solutions in respective fields. 

For ant tracking, we have only a few publicly available datasets~\cite{naturemethodsanttracking,antdataset1}. The dataset from~\cite{antdataset1} is relatively small with about $5k$ images. The dataset published in~\cite{naturemethodsanttracking} has $20K$ annotated images but only sparse object detections with $1.8$ average detections per frame. Due to this sparseness we cannot observe complex motions and frequently overlapping tracks. Both of these datasets contain videos from only one ant species. Our   proposed large-scale dataset  has two ant species and higher density of $25.6$ detections per frames allowing us to assess complex tracking scenarios. Furthermore, we establish standard benchmark tasks for this dataset.

\section{Framework}
\begin{figure}[tb]
\begin{center}
\includegraphics[width=0.5\textwidth]{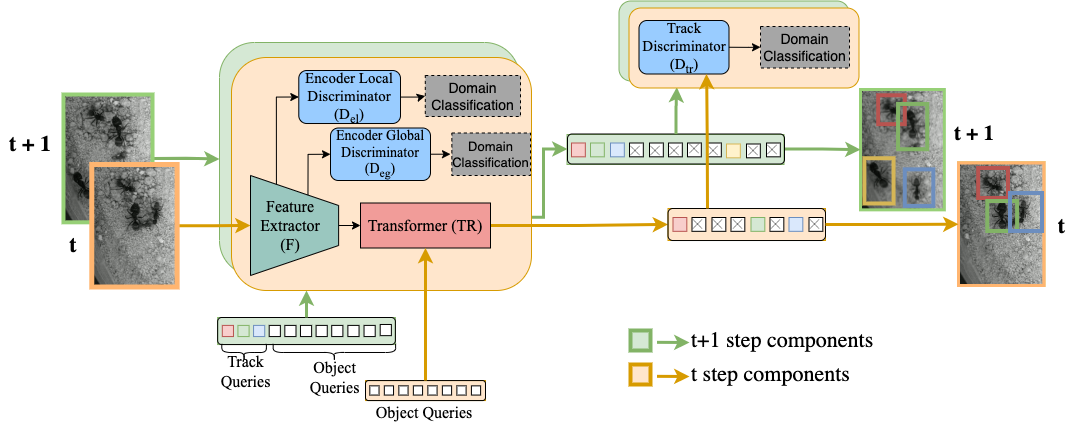}
\end{center}
   \caption{A schematic of our proposed network architecture. The multi-object tracking module predicts detected boxes and associate them between two frames in a unified network. To ensure our proposed framework can adapt to the target data distribution, the intermediate layer feature representations are enforced to follow a common data distribution. The conversion happens at two levels: Image encoded features( $D_{eg}$ and $D_{el}$), track level ($D_{tr}$).}
\label{fig:networkdesign}
\end{figure}

We present an unsupervised domain adaptation method for a multi-object tracking network that translates source and target image features into a common distribution. Mapping the target feature distribution to the source distribution, we can expect a tracker trained only on the source domain tracker to retain its performance in the target domain. Our proposed approach relies on several adversarial losses for classifying image-level and track-level features between source and target data. As depicted in \Fig\ref{fig:networkdesign}, our adversarial approach relies on three discriminators (blue boxes): two of these operate on the encoded visual features, while the third one operates on the decoded features representing  the outputs/trajectories.

\subsection{Multi-object tracker}
Our approach is based on the recently proposed end-to-end multi-object tracker, known as Trackformer~\cite{trackformer}. In this framework, the tracker component uses two extended object detectors that work on consecutive frames and links the individual predictions using \emph{track queries}. These track queries are essentially a summary of information about object detections in the previous frame that is passed to a predictor for the current frame. To train the tracker on the source domain, we use the supervised loss, $\mathcal{L}_{MOT,S}$, suggested by~\cite{trackformer}, which is a combination of losses for track initiation and localization for new objects and state prediction for the existing ones.

\subsection{Domain adaptation in layers}

To extend this tracking framework for unsupervised domain-adaption, we adopt the adversarial training strategy by integrating three different discriminators, trained simultaneously with the multi-object tracker. These discriminators enforce similarity between the encoded/decoded features from source and domain data. Their details are elaborated below:

\textbf{Encoded visual feature alignment: } Feature alignment works at frame-level feature extraction layers in the backbone $F$. One discriminator ($D_{el}$ in Figure \ref{fig:networkdesign}) operates on local features, the other one ($D_{eg}$) on  global features (\eg\ backgrounds, scene layouts). At the global level, domain-specific attributes may vary significantly, and attempts to align them may negatively impact model performance~\cite{strongweakfeaturealignment}. By incorporating two discriminators, the degree of feature alignment can be tailored to the specific layer.

$D_{el}$ takes its input of height $H$ and width $W$ from the backbone layer $(F')$ and produces a domain classification prediction of the same shape. A pixel level loss is estimated from this prediction for each source domain frame, $x^S_i$, and target domain frame, $x^T_i$, to establish a strong alignment cost for local features. This local feature classification cost, $\mathcal{L}_{local}$ is calculated over two consecutive time steps, $t=1,2$:

\begin{equation}
    \mathcal{L}_{loc_{S,t}} = \frac{1}{n_SH\,W}\sum_{i=1}^{n_S}D_{el}(F'(x_{i,t}^S))^2 
\end{equation}
\begin{equation}
    \mathcal{L}_{loc_{T,t}} = \frac{1}{n_TH\,W}\sum_{i=1}^{n_T}(1-D_{el}(F'(x_{i,t}^T)))^2
\end{equation}
\begin{equation} \label{eq:strongequation}
    \mathcal{L}_{local} = \frac{1}{4}\sum_{t=1,2}\mathcal{L}_{loc_{S,t}}+\mathcal{L}_{loc_{T,t}}
\end{equation}

$D_{eg}$ similarly predicts the domain category at the global feature level (Equation \ref{eq:weakequation}). This loss, a modified version of cross-entropy loss, assigns a higher value for hard-to-classify examples~\cite{strongweakfeaturealignment}.

\begin{align}
    \mathcal{L}_{gl_{S,t}} = -\frac{1}{n_S}\sum_{i=1}^{n_S}(1-&D_{eg}(F''(x_{i,t}^S))^\gamma\nonumber \\&\times\log(D_{eg}(F_2(x_{i,t}^s)))
\end{align}
\begin{align}
    \mathcal{L}_{gl_{T,t}} = -\frac{1}{n_T}\sum_{i=1}^{n_T}&D_{eg}(F''(x_{i,t}^T)^\gamma \nonumber \\&\times\log(1 - D_{eg}(F''(x_{i,t}^T)))
\end{align}
\begin{equation} \label{eq:weakequation}
    \mathcal{L}_{global} = \frac{1}{4}\sum_{t=1,2}\mathcal{L}_{gl_{S,t}}+\mathcal{L}_{gl_{T,t}}
\end{equation}

Our ablation experiments provided in Sec.~\ref{sec:exp} show that the visual feature alignment by itself already yields a significant improvement in the tracking performance for the target domain data, compared to the supervised baseline. This performance improvement is mainly due to better object detection in the target domain data. To compensate domain shift between the outputs (tracks), we also add a track-level alignment component by an additional track discriminator.

\textbf{Track level alignment:} 
We introduce a track discriminator, $D_{tr}$ at the final output layer of the Transformer ($TR$) to compensate the domain shift between the output (track) distributions. The output from the transformer, $q_{i,j}$ contains an embedding of both track queries and object queries. These are used as the initialisation for the continuation of the track. Domain alignment at the track level thus addresses track continuation and track termination.

The track discriminator, $D_{tr}$, ensures that the decoded (output-level) features from the source and domain data are not distinguishable by classifying their domain individually in an adversarial learning setting 
using the following losses:
\begin{equation} 
\begin{split}
    \mathcal{L}_{tr_{S,t}} = \frac{1}{n_S(n_{tr}+n_{ob})}\sum_{i=1}^{n_S}\sum_{j=1}^{n_{tr}+n_{ob}}(1-D_{tr}(q_{i,j}))^\gamma \\ \times \log(D_{tr}(q_{i,j}))
\end{split}
\end{equation}
\begin{equation}
\begin{split}
    \mathcal{L}_{tr_{T,t}} = \frac{1}{n_T(n_{tr}+n_{ob})}\sum_{i=1}^{n_T}\sum_{j=1}^{n_{tr}+n_{ob}}(D_{tr}(q_{i,j}))^\gamma \\ \times \log(1-D_{tr}(q_{i,j}))
\end{split}
\end{equation}
\begin{equation} \label{eq:trackcostfucntion} 
    \mathcal{L}_{track} = \frac{1}{4}\sum_{t=1,2}\mathcal{L}_{tr_{S,t}}+\mathcal{L}_{tr_{T,t}}
\end{equation}
\subsection{End-to-end trainable domain adaptation}

Finally, Equation \ref{eq:totallossfunction} combines the loss of all three discriminators with the supervised tracking loss, $L_{MOT, S}$. To train the model with all the losses together in an end-to-end trainable fashion we integrate a gradient reverse layer~\cite{grl} between the discriminators and the tracker. The gradient reverse layer introduces negative feedback when the source and target domains features are easily distinguishable. The rationale is that easy discrimination can be assumed to be detrimental to the domain transfer of tracking. In this way the model can be trained for multiple objectives simultaneously.

\begin{equation} \label{eq:totallossfunction}
    \mathcal{L}_{total} = \mathcal{L}_{MOT, S} + \lambda_1\mathcal{L}_{local} + \lambda_2\mathcal{L}_{global} + \lambda_3\mathcal{L}_{track}
\end{equation}

\section{Dataset and Benchmark}

\begin{figure*}[t]
\centering

\tabskip=0pt
\valign{#\cr
  \hbox{%
    \begin{subfigure}[b]{.38\textwidth}
    \centering
    \includegraphics[height=6.1cm, width=\textwidth]{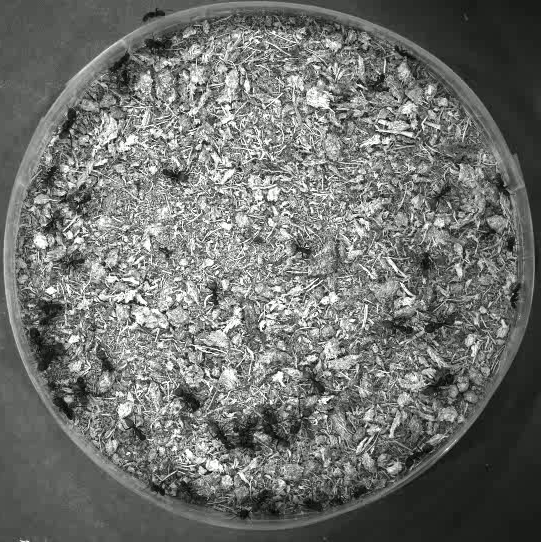}
    \caption{Highly crowded target domain sample}
    \label{samples.a}
    \end{subfigure}%
  }\cr
  \noalign{\hfill}
  \hbox{%
    \begin{subfigure}{.58\textwidth}
    \centering
    \includegraphics[width=.8\textwidth]{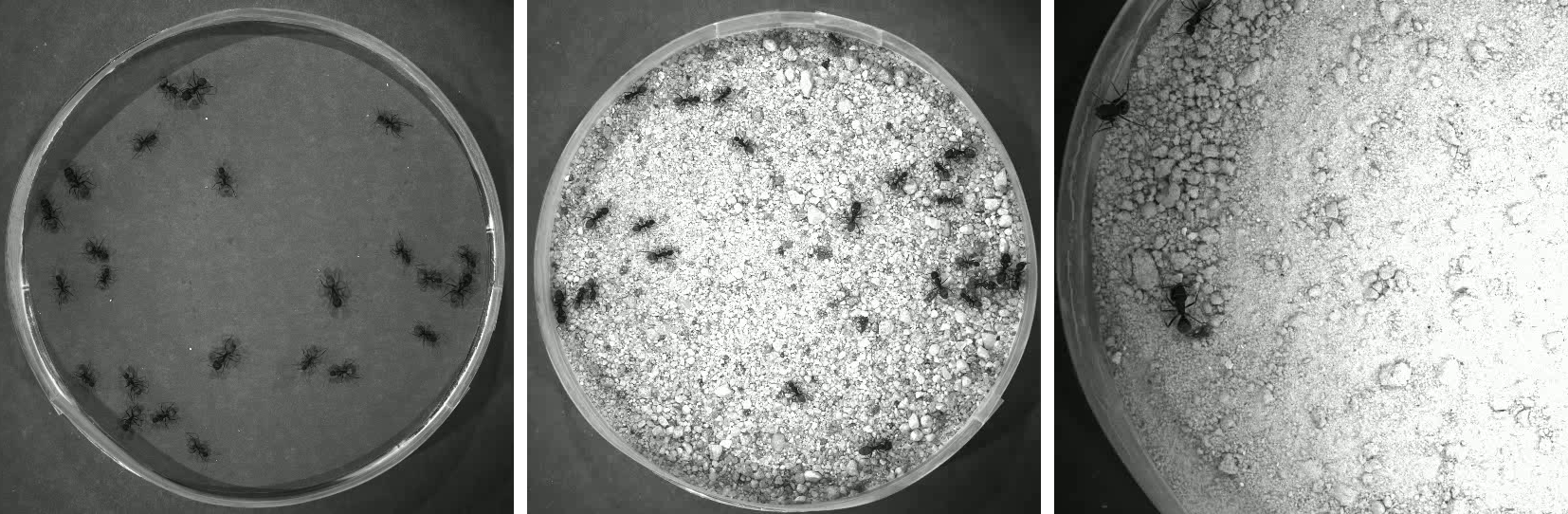}
    \caption{Target domain samples}
    \label{samples.b}
    \end{subfigure}%
  }\vfill
  \hbox{%
    \begin{subfigure}{.58\textwidth}
    \centering
    \includegraphics[width=.8\textwidth]{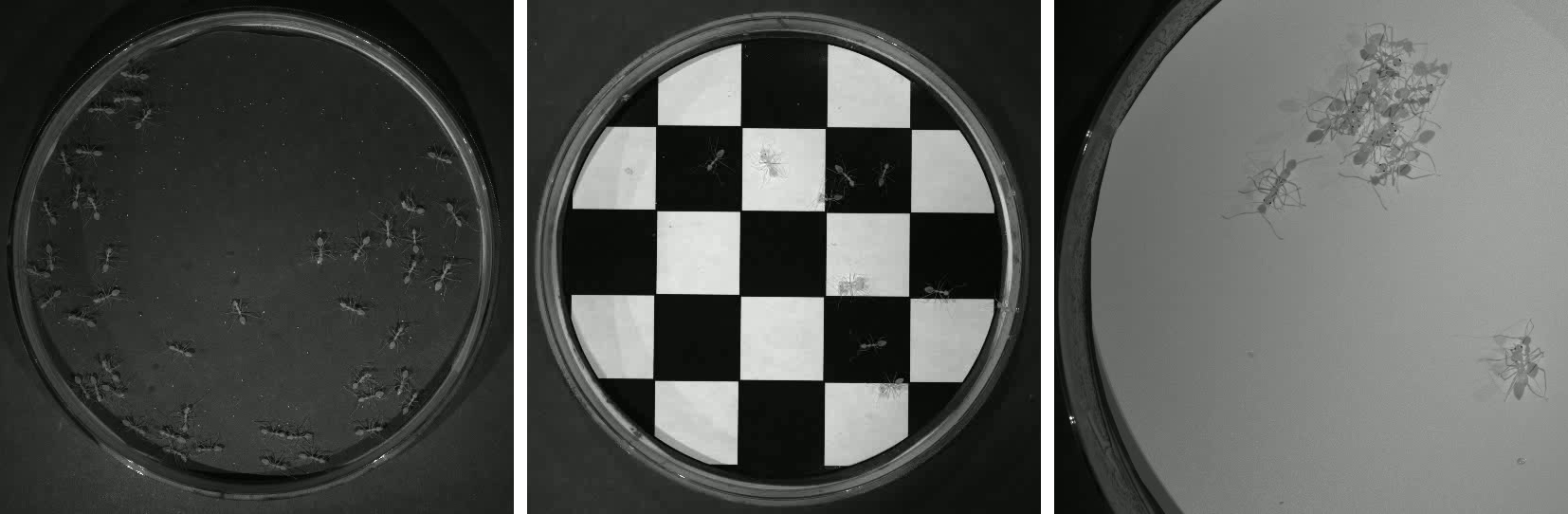}
    \caption{Source domain samples}
    \label{samples.c}
    \end{subfigure}%
  }\cr
}

\caption{Samples taken from our dataset to show the diversity. Dataset has two ant species divided as target domain and source domain. Target domain has four videos in four different backgrounds and source domain has 3 different backgrounds.}
\label{fig:data-examples}

\end{figure*}

\subsection{Annotated Dataset}

\begin{table}[b]
  \begin{center}     
    {\scriptsize{}{
\setlength\tabcolsep{2.5pt}
\begin{tabular}{l|rrrr|rrrr}
\toprule

  & Source  &  &  & & Target  &  & &  \\
 \toprule

  & Train  & Val & Test & All & Train  & Val & Test & All \\
\midrule
Tot. videos & 20 & 6 & 7 & 33 & 12 & 6 & 6 & 24 \\
Tot. frames & 12200 & 3440 & 4160 & 19800 & 8200 & 4100 & 4100 & 16400 \\
Tot. detections & $286K$ & 82K & 103K & 471K & 135K & 53K & 60K & 249K \\
Tot. tracks & 635 & 217 & 199 & 1051 & 455 & 215 & 203 & 873 \\
Avg. track len. & 450.0 & 379.8 & 518.7 & 448.5 & 297.0 & 249.8 & 298.5 & 285.7 \\
\bottomrule
\end{tabular}
}}
\end{center}
\caption{Our dataset has two parts: Source and Target. For domain adaptation applications, we assume we have labels only for source domain data during training. To establish a benchmark, we divide both source and the target data into training, validation and test splits.  }
\label{table:datasetsummary1}
\end{table}

\begin{figure}[t]
\begin{center}
    
\begin{subfigure}[t]{0.35\linewidth}
    \centering
         \includegraphics[width=\linewidth]{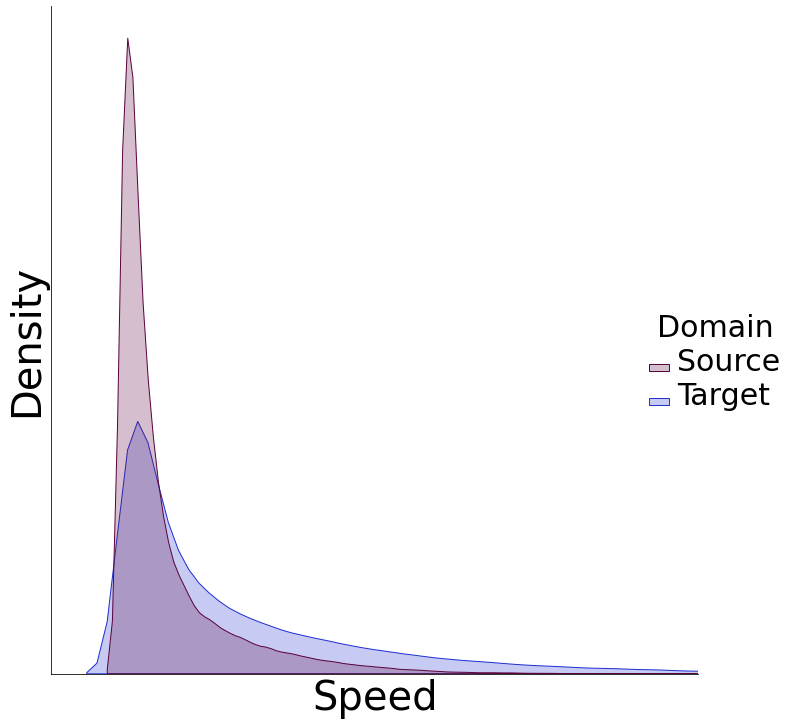}
         \caption{Speed distribution}
         \label{fig:speeddistro}
\end{subfigure}
\begin{subfigure}[t]{0.35\linewidth}
    \centering
         \includegraphics[width=\linewidth]{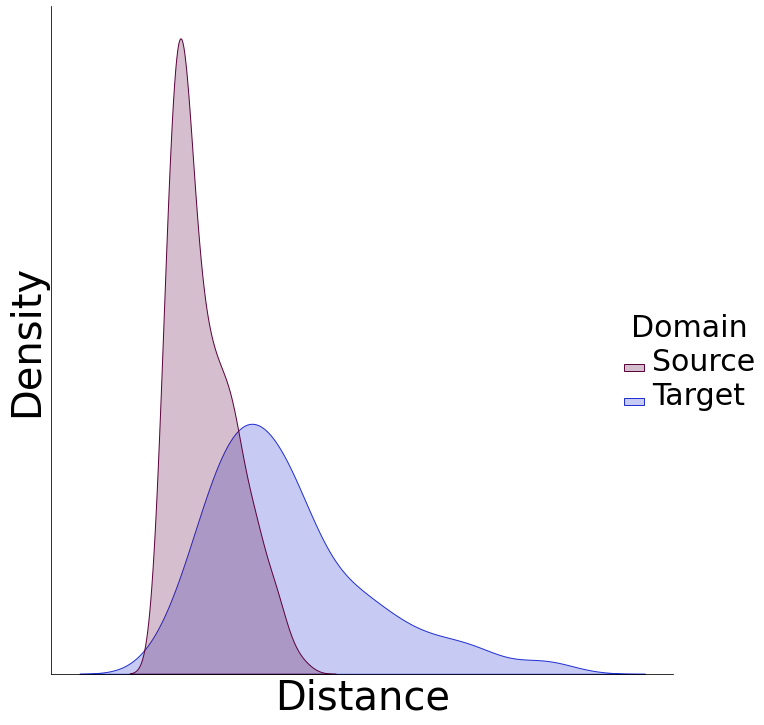}
         \caption{Travel distance distribution}
         \label{fig:distancedistro}
\end{subfigure}

\begin{subfigure}[t]{0.35\linewidth}
    \centering
         \includegraphics[width=\linewidth]{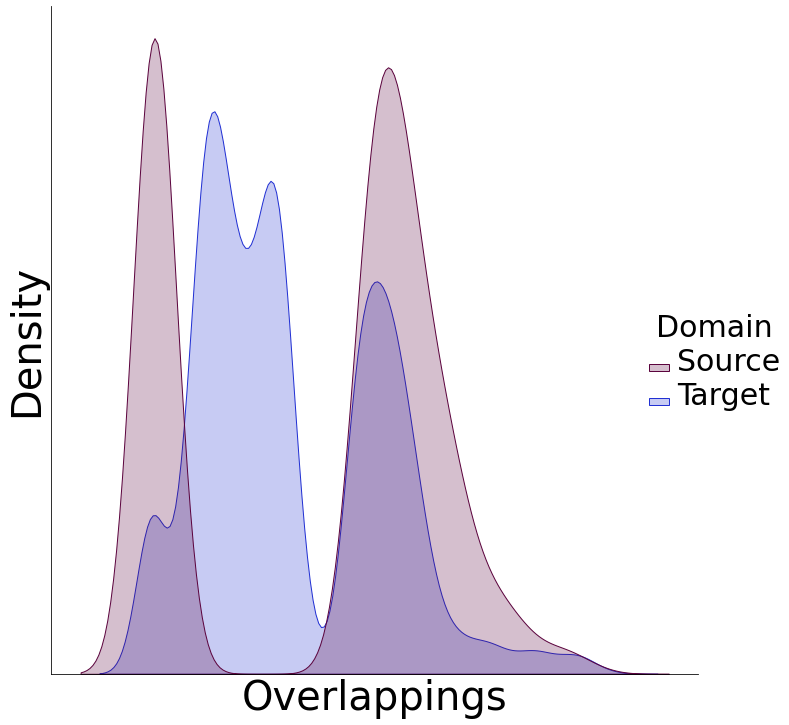}
         \caption{Crowd density distribution}
         \label{fig:crowddistro}
\end{subfigure}
\begin{subfigure}[t]{0.35\linewidth}
    \centering
         \includegraphics[width=\linewidth]{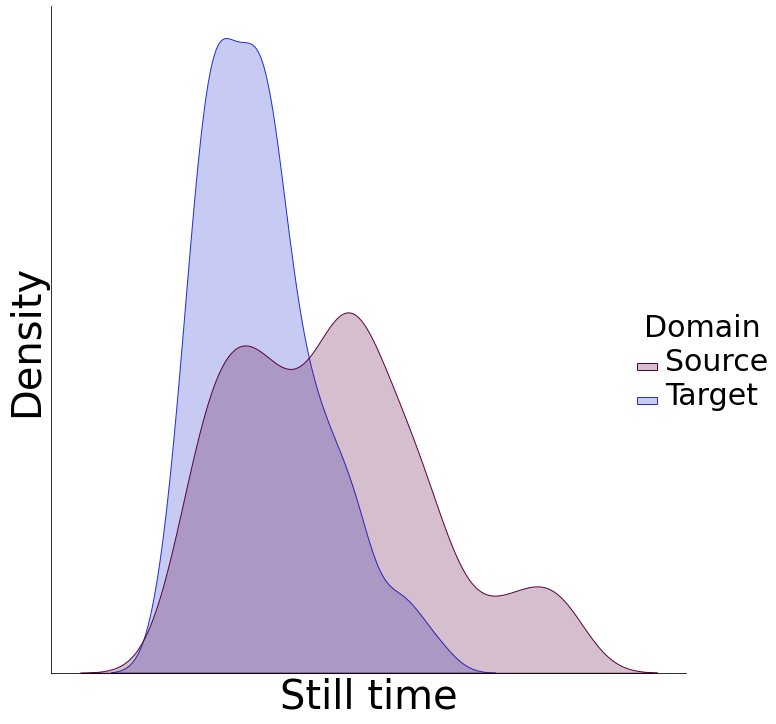}
         \caption{No-moving time distribution}
         \label{fig:stilltimedistro}
\end{subfigure}

\end{center}

   \caption{Distribution difference in target and source domain. Here we illustrate behavioural pattern difference between two ant species. }

\label{fig:dataset_distribution}
\end{figure}

Our benchmark dataset contains 57 video sequences recorded in a standard laboratory environment in the context of a real biological experiment. It comprises  33 videos in the source domain and 24 videos in the target domain for a total of 36k frames and 700k object detections. Each video contains between 10-50 ants on average. The videos were recorded with different backgrounds, zoom levels and lighting conditions. Table~\ref{table:datasetsummary1} details the dataset metrics. 
Our dataset comprises videos of two very different species of ants: Weaver ants (\emph{Oecophylla smaragdina}) and Carpenter ants (\emph{Camponotus aeneopilosus})~\cite{wilsontheants}. 
As shown in Figs.~\ref{fig:data-examples} and~\ref{fig:dataset_distribution}, these two ant species have clearly distinguishable physical appearances and behavioural patterns, making this dataset a proper practical benchmark for domain adaptation in multi-object tracking.

We produced the ground-truth track using a two-step process: (1) a simple tracking algorithm was used to produce a tentative, partial and noisy ground truth (2) all these initial annotations were manually corrected by a human expert. 

For the initial phase, we first trained a Faster R-CNN detector~\cite{fasterrcnn} using the detected ant locations by an off-the-shelf QR tracker~\cite{qrtracking} on an external dataset including ants tagged with micro-QR codes.  
We then used this trained detector on a second set of videos including untagged ants, \ie our final benchmark dataset. 
Next, a basic multi-object tracking approach (Kalman filtering followed by Hungarian matching), was applied to link detections between successive frames and to produce the initial track annotations. Due to the fact that the object detector was trained for QR-tagged ants and due to the simplistic tracking linking, this ``pseudo ground-truth'' is obviously not useable as a real ground-truth. 
Thus, in the second phase, the ``pseudo ground-truth'' was manually and carefully corrected by human inspection to generate a fully valid ground-truth.

\begin{table*}[h]
  \begin{center}
    {\scriptsize{}{
\begin{tabular}{llllllllll}
\toprule
Detector & Tracker & MOTA & IDF1 & HOTA & MT & ML & Frag & FP(/frame) & ID Sw. \\
\midrule

DETR~\cite{detr}$^\dagger$ & SORT~\cite{sortalgorithm} & -3.759 & 0.107 & 0.203 & 54 & 69 & 673 & 61.75 & 582 \\
DETR~\cite{detr}$^\dagger$  & EMD tracker~\cite{naturemethodsanttracking}  & -3.830 & 0.110 & 0.205 & 61 & 65 & 731 & 63.20 & 640 \\
\multicolumn{2}{c}{Transtrack~\cite{transtrack}$^\dagger$ (joint detection and tracking )} & -3.606 & 0.066 & 0.132 & 60 & 84 & 3210 & 59.76 & 2546\\
\multicolumn{2}{c}{Trackformer~\cite{trackformer}$^\dagger$ (joint detection and tracking )} & -1.02 & 0.187 & 0.262 & 57 & 118 & 1148 & 20.87 & 1351 \\

 &  &  &  &  &  &  &  &  & \\

\hline

DA FRCNN~\cite{strongweakfeaturealignment}$^\ddagger$ & SORT~\cite{sortalgorithm}  & 0.216 & 0.303 & 0.288 & 38 & 57 & 1874 & 4.53 & 1086 \\
DA FRCNN~\cite{strongweakfeaturealignment}$^\ddagger$  & EMD tracker~\cite{naturemethodsanttracking}  & 0.225 & 0.333 & 0.316 & 52 & 31 & 3402 & 5.79 & 1223 \\
 &  &  &  &  &  &  &  &  & \\

\hline

\textbf{Adapt DETR}$^\ddagger$ & SORT~\cite{sortalgorithm}  & 0.411 & 0.254 & 0.284 & 96 & 40 & 1851 & 2.70 & 1723 \\ 
\textbf{Adapt DETR}$^\ddagger$ &  EMD tracker~\cite{naturemethodsanttracking}  & 0.413 & 0.275 & 0.305 & \textbf{112} & \textbf{21} & 2513 & 4.05 & 2085 \\
\multicolumn{2}{c}{\textbf{DA-Tracker}$^\ddagger$ (joint detection and tracking )} & \textbf{0.493} & \textbf{0.494} & \textbf{0.433} & 42 & 42 & \textbf{755} & \textbf{1.261} & \textbf{495} \\
 &  &  &  &  &  &  &  &  &  \\
\hline
\hline
\multicolumn{2}{c}{Transtrack~\cite{transtrack}$^{\dagger\dagger}$ (joint detection and tracking )} & 0.918 & 0.584 & 0.584 & 200 & 1 & 819 & 0.491 & 823  \\

\multicolumn{2}{c}{Trackformer~\cite{trackformer}$^{\dagger\dagger}$ (joint detection and tracking )}  & 0.912 & 0.646 & 0.610 & 180 & 4 & 615 & 0.309 & 463 \\

\bottomrule
\end{tabular}
}}
\end{center}

\caption{
Comparison of multi-object trackers on the target domain with and without domain adaptation techniques. We report three sets of baselines. In the first set we train non-adaptive multi-object trackers only on source domain data and evaluate on target data.  In the second set of baselines, we do the same with domain-adaptive object trackers. Finally, as an upper bound, we evaluate fully supervised trackers directly trained on the target domain data. Highlighted methods are our contributions. $\dagger$ - Trained on source domain labels and images $\ddagger$ - Trained on source domain labels, images and target domain images   $\dagger\dagger$ - Trained on target domain labels and images. 
}
\label{comparisontable}

\end{table*}

\subsection{Benchmark}

\textbf{Source and target domains: }
The main application of our benchmark is to assess multi-object tracking frameworks in an unsupervised domain adaptation experimental setting. For this benchmark, we consider the weaver ant videos as the source domain 
and the carpenter ant videos as the target domain data, respectively.


\textbf{Train, validation and test split:} We divided both the target and
source domain data into three splits: training, validation and
test. The training split comprises about 50\% of data; test
and validation about 25\% each. 
We tried our best to ensure the statistics of each split reflects the similar distribution in the terms of background types, zoom levels, density of ants \emph{etc}. We will use the training and validation sets of the source (inputs and their annotations) and target domains (inputs only) for optimizing the model parameters and hyper-parameter tuning, respectively. The final results are evaluated on the target domain's test split. 

\textbf{Evaluation metric:} Tracking performances in the biological/ecological literature are reported in a variety of different ways that do not allow direct comparison. To overcome this, we adopt standards of the multi-object tracking literature: MOTA, HOTA and IDF1. In addition to these, we include other key metrics like IDSW, MT, ML in the benchmark. An explanation of the full details and intricacies of these metrics is beyond the scope of this paper. Full details are given in \cite{hotametric,idfmetric,clearmotmetric}. In broad sketch terms, tracking requires so solve two sub-tasks (object localization and detection association) and these metrics emphasize the performance on these two sub-tasks to different extents.  IDF1 is biased toward tracking accuracy, while  MOTA  is biased toward detection accuracy.  HOTA provides an overall assessment of both of these factors. 

\begin{figure*}[ht]
\begin{center}
    
\begin{subfigure}[t]{0.26\linewidth}
    \centering
         \includegraphics[width=\linewidth]{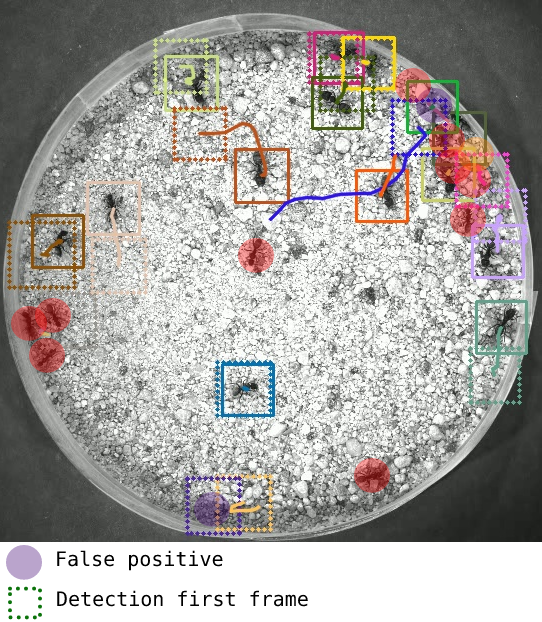}
         \caption{DA-FRCNN and SORT baseline}
         \label{fig:tracks_baseline}
\end{subfigure}
\begin{subfigure}[t]{0.26\linewidth}
    \centering
         \includegraphics[width=\linewidth]{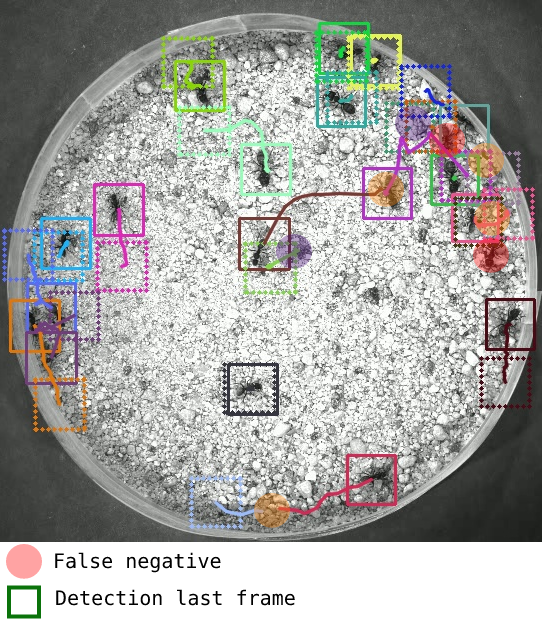}
         \caption{Our approach}
         \label{fig:tracks_ours}
\end{subfigure}
\begin{subfigure}[t]{0.26\linewidth}
    \centering
         \includegraphics[width=\linewidth]{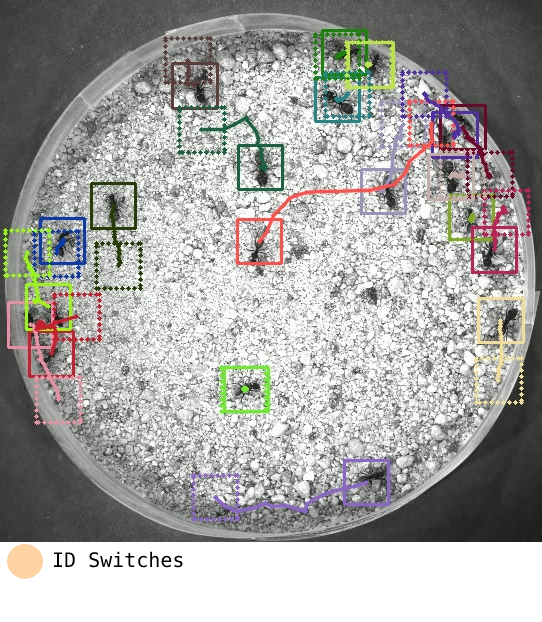}
         \caption{Ground truth}
         \label{fig:tracks_gt}
\end{subfigure}

\end{center}
\vspace{-1em}

\caption{Visualization of tracks between for twenty consecutive frames. Domain adaptive detection based tracker in \ref{fig:tracks_baseline} has many incorrect predictions in crowded areas compared to our proposed approach. }
\label{fig:track_visualize}
\end{figure*}

\begin{figure}[ht]

\begin{center}
\begin{subfigure}[t]{0.42\linewidth}
    \centering
         \includegraphics[width=.95\linewidth]{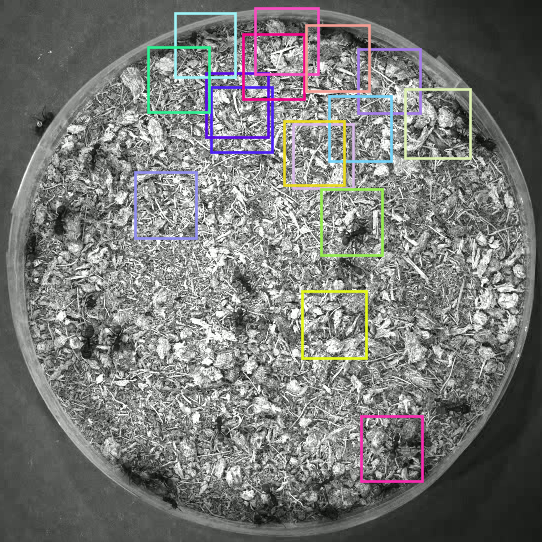}
         \caption{Trackformer trained on source data}
         \label{fig:tracks_baseline_detections}
\end{subfigure}
\begin{subfigure}[t]{0.42\linewidth}
    \centering
         \includegraphics[width=.95\linewidth]{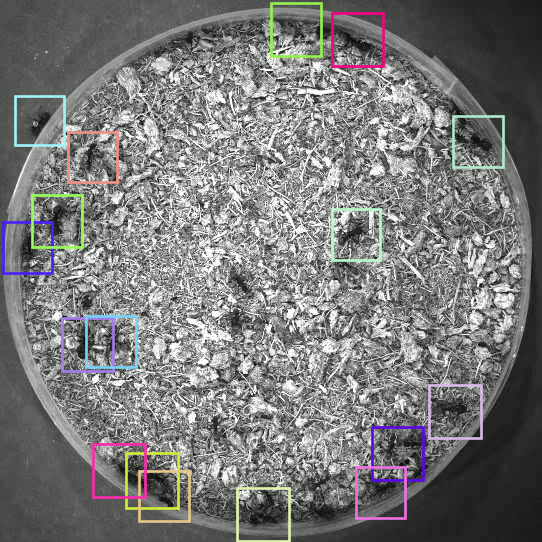}
         \caption{Our approach}
         \label{fig:tracks_ours_detections}
\end{subfigure}
\end{center}
\vspace{-1em}
\caption{We achieved a significant tracking improvement by applying our proposed domain adaptation techniques. This illustrates detections before applying domain adaptation (a) and after applying domain adaptation (b).}
\label{fig:predictionsamples}
\end{figure}

\section{Experiments}\label{sec:exp}


\textbf{Implementation} 
Our proposed architecture incorporates a multi-object tracker that is connected to a set of discriminators at different levels. The MOT component of this network is based on \cite{trackformer}, which is a joint-detector-tracker that has a feature extraction backbone and a transformer encoder-decoder network. Our method employs ResNet101 as the backbone for feature extraction. The transformer network, which takes both the backbone output and a positional embedding as input, is composed of a 6-layer encoder and a 6-layer decoder. This network includes three discriminators: two image feature discriminators and a track discriminator. The image feature discriminators are convolutional networks that use the output from the first and third layers of the backbone as local-level and global-level input features, respectively. They have three convolutional layers each. The track discriminator is an Multi-layer perceptron with two fully connected layers. It classifies the output embedding from the transformer decoder. We use gradient reverse layers to integrate the discriminators. Full details of the implementation are available in supplementary materials.

\textbf{Experimental Evaluation}
We compare our method with a selection of representative baseline models on our dataset. We train our unsupervised domain adaptation MOT model on the training set of the source domain using the provided annotations and the training set of target domain (images only). Then we evaluated the model on the target domain test split. 

For a direct comparison with our proposed domain adaptation tracking method, we did not find any relevant literature based on deep learning to address the domain adaptation for MOT problem directly. Therefore, we compare detection-based trackers that use domain adaptation methods in the object detection step. DA-FRCNN~\cite{strongweakfeaturealignment} is a version of Faster R-CNN which uses domain adaptation techniques. We create tracks from these detections by using SORT~\cite{sortalgorithm} (Kalman filtering and Hungarian matching) and an earth-mover distance-based~\cite{naturemethodsanttracking} trackers. We produce an additional not previously published baseline by modifying  DETR~\cite{detr} to perform domain adaptation using two discriminators working on image-level features similar to our proposed approach. This domain adaptive detection method is named as ``Adapt DETR'' in our experiments. We also use SORT and EMD to generate tracks using this domain adaptive detector.

For completeness, we also compare our proposed approach to baselines that do not use domain adaptation. These are  Trackformer~\cite{trackformer}, and standard DETR~\cite{detr} with tracking using SORT~\cite{sortalgorithm} and EMD~\cite{naturemethodsanttracking}. The latter approach is closest to what has recently been proposed as a state-of-the-art tracker for real-world ant tracking in the biological literature \cite{naturemethodsanttracking}.


\Tab \ref{comparisontable} summarizes our  results. To measure the improvement achieved by applying domain adaptation, we compare against the models that do not use domain adaptation. Compared to Trackformer, we achieve more than $40\%$ improvement in HOTA, MOTA and IDF1 metrics. \Fig~\ref{fig:predictionsamples} illustrates the difference between predictions in these two experiments. We note similar significant improvement against other non-domain adaptive approaches.

Compared with detection-based trackers that use domain adaptation in the detection component alone, our methods shows more than $27\%$ improvement in HOTA, MOTA and IDF1. Compared to the best performing baseline, Adapt DETR, \ie DETR with integrated domain adaptation strategy, followed by SORT or EMD tracker, our method still shows more than $16\%$ improvement. \Fig~\ref{fig:track_visualize} shows the difference between baseline and proposed approach track predictions.

To report an upper bound on the performance, \ie\ an overly optimistic best case scenario, we trained a separate model directly on the target domain's training split in a supervised way assuming the annotated data is available. As expected, this approach performs better than our proposed framework.

\textbf{Ablation study} 
Table~\ref{table:results2} shows the result of the ablation study conducted to understand the contribution of each component.
The results demonstrate the each discriminator's contribution to the final performance. Using only the visual feature discriminators, the performance can be improved considerably; However it still produces many false positives. This higher false positive rate is a result of object queries and track queries not being properly optimised for the target domain data. The integration of the track-level discriminator significantly reduces the number of false positives. 

\begin{table}[h]
  \begin{center}
    {\scriptsize{}{
    \setlength\tabcolsep{2.5pt}
\begin{tabular}{cccccccccccc}
\toprule
$D_{el}$ & $D_{eg}$ & $D_{tr}$ &  MOTA & IDF1 & HOTA & MT & ML & Frag & FP & ID Sw. \\
\midrule
\textendash & \textendash & \textendash & -1.02 & 0.187 & 0.262 & 57 & 118 & 1148 & 20.87 & 1351 \\
 \Checkmark & \textendash & \textendash &  0.160 & 0.305 & 0.365 & 70 & 88 & 1916 & 5.021 & 5684 \\
 \textendash & \Checkmark & \textendash & 	0.301 &	0.325 &	0.331 &65	& 48 &	1621 &	4.625 &	1996\\
 \Checkmark & \Checkmark & \textendash &  0.381 & 0.453 & \textbf{0.445} & \textbf{118} & \textbf{10} & 1089 & 5.406 & 880\\
 \Checkmark & \Checkmark & \Checkmark &  \textbf{0.493} & \textbf{0.494} & 0.433 & 42 & 42 & \textbf{755} & \textbf{1.261} & \textbf{495}\\

\bottomrule
\end{tabular}
}}
\end{center}
\caption{We conduct an ablation study to evaluate the accuracy improvement of domain adaptation components: Encoded feature local discriminator$(D_{el})$, Encoded feature global discriminator$(D_{eg})$ and Track discriminator$(D_{tr})$. Combining $D_{el}$ and $D_{eg}$ yields significant improvement compared to individual components, but it results in many false detections due to incorrect predictions propagating to future predictions. $D_{tr}$ address this issue by converting transformer decoder output embeddings.}
\label{table:results2}

\end{table}

\section{Conclusion}
We have introduced a new multi-object tracking model capable of domain transfer. This was achieved by integrating multiple unsupervised adversarial discriminators at different processing stages into a joint-detection-and-tracking model. Our experiments have shown that our approach can achieve noticeable performance improvements when tracking objects in a new target domain data with different  visual appearances and shifted output distributions. This is exactly the case for many ant experiments in collective behaviour studies that use different ant species and different experimental assays.

Multi-object tracking of ant experiments is a core component of experiments in collective behaviour research but training a new model for every different setup and species is not practically feasible, since generating training data is as laborious and costly as tracking the experiment manually and in some cases even more. It is mostly for this reason that many experiments have to fall back on  manual tracking. This severely limits experiment sizes and  data throughput  and thus  the value and reach of such experiments. Our technique can clearly help to alleviate this bottleneck.

To encourage other researchers to work in the same space, we have compiled and provided a new benchmark dataset for ant tracking based on realistic laboratory experiments. 

It is clear that this is only the beginning of making untagged multi-insect tracking a routine component in such experiments. We are continuing to expand this proposed model and specifically, we  plan to next address the transfer between multiple domains. This will be reflected in the public benchmark dataset that will be extended accordingly.

\newpage

\bibliographystyle{named}
\bibliography{main}

\end{document}


\beginsupplement
\maketitle

\section*{Introduction}
This document serves as a supplement to our paper.  It includes details on the implementation and training of our proposed method, DA-Tracker, as well as implementation specifics for Adapt DETR, one of the baselines used in our study. Additionally, we have included an example video that demonstrates our results.

\section*{DA-Tracker implementation}

DA-Tracker is a combination of two key components: a Multi-Object Tracker (MOT) and domain adaptive discriminator modules. The MOT component of DA-Tracker utilizes a Trackformer network\footnote{We use the original trackformer github code available at https://github.com/timmeinhardt/trackformer}, which utilizes a ResNet101 backbone as its underlying architecture. To enhance the performance of the MOT module, we have integrated three discriminator modules: two image feature discriminators and one track discriminator. These discriminator modules are connected to different parts of the Trackformer network. The first and third layers of the ResNet backbone are connected to the two image feature discriminators, while the track discriminator is connected to the output of the Transformer decoder. The image feature discriminators are implemented using a fully convolutional neural network (FCNN), while the track discriminator is implemented as a fully connected network (FCN).

Tab. \ref{tab:discriminator1}, Tab. \ref{tab:discriminator2} and Tab. \ref{tab:discriminator3} shows the layer structure for our local discriminator, global discriminator and track discriminator respectively. 

\begin{table}[h]
    \centering
    \begin{tabular}{lrr}
        \toprule
        Layer  & Kernel  & Stride \\
        \midrule
        Conv2D & $256\times(1\times1)$ & $(1\times1)$ \\
        Conv2D & $128\times(1\times1)$ & $(1\times1)$ \\ 
        Conv2D & $1\times(1\times1)$ & $(1\times1)$ \\
        \bottomrule
    \end{tabular}
    \caption{Pixel-wise local image feature discriminator layers}
    \label{tab:discriminator1}
\end{table}

\begin{table}[h]
    \centering
    \begin{tabular}{lrr}
        \toprule
        Layer  & Kernel  & Stride \\
        \midrule
        Conv2D & $512\times(3\times3)$ & $(2\times2)$ \\
        BatchNormal2D & - & - \\
        Dropout (p=0.5) & - & - \\
        Conv2D & $128\times(3\times3)$ & $(2\times2)$ \\ 
        BatchNormal2D & - & - \\
        Dropout (p=0.5) & - & - \\
        Conv2D & $128\times(1\times1)$ & $(2\times2)$ \\
        BatchNormal2D & - & - \\
        Dropout (p=0.5) & - & - \\
        AveragePool2D & - & - \\
        Linear(128) & - & - \\
        
        \bottomrule
    \end{tabular}
    \caption{Global image feature discriminator layers}
    \label{tab:discriminator2}
\end{table}

\begin{table}[h]
    \centering
    \begin{tabular}{lrr}
        \toprule
        Layer  & Units \\
        \midrule
        Linear & 128  \\
        Dropout (p=0.5) & - \\
        Linear & 2 \\ 
        \bottomrule
    \end{tabular}
    \caption{Track discriminator layers}
    \label{tab:discriminator3}
\end{table}

\section*{Training procedure and optimization}
During the training, we trained the DA-Tracker for 30 epochs using a specific approach. The first 5 epochs focused on training track initialization by setting the false track query probability parameter to 1.0. This removes all previous detections from track query and forces the detector to initialize tracks for each frame. After that, we trained for the next 5 epochs with the default hyper-parameters. Then, for epochs 10-15, we increased the false positive probability to 0.8 to train track termination. Finally, we trained with the default parameters again. Additionally, during training, the $\gamma$ value in our weak global cost function was set as 2.0. 
We set discriminator loss coefficients ($\lambda_1, \lambda_2, \lambda_3$) as 100.0. We used the AdamW optimizer and batches of two samples for training. The model is trained on a NVIDIA Tesla T4 GPU with 16GB memory.

\section*{Adapt-DETR implementation}

In our implementation of Adapt DETR, we made modifications to the DETR network\footnote{We use the original DETR GitHub code available at https://github.com/facebookresearch/detr} by incorporating domain adaptation modules. Similar to our multi-object tracker, we used two discriminators connected to the ResNet backbone to convert image-level features. The two discriminators are similar to the image-level discriminators in the DA-Tracker, and have the same layer structure as outlined in Tab. \ref{tab:discriminator1} and Tab. \ref{tab:discriminator2}. To connect the predictions from Adapt DETR, we employed the SORT and EMD algorithms.

\begin{table}[t]
  \begin{center}
    {\scriptsize{}{
    \setlength\tabcolsep{2pt}
\begin{tabular}{lrrrrrrrr}
\toprule
Sequence & MOTA & IDF1 & HOTA & MT & ML & Frag & FP(/frame) & ID Sw. \\
\midrule
DETR + SORT &  &  &  &  &  &  &  & \\ 
CU10L1B6In & -25.004 & 0.006 & 0.016 & 0 & 9 & 235 & 98.77 & 166\\ 
CU15L1B1In & 0.924 & 0.792 & 0.736 & 33 & 1 & 50 & 0.36 & 44\\ 
CU15L1B4In & -11.589 & 0.039 & 0.082 & 1 & 5 & 289 & 93.94 & 299\\ 
CU20L1B1Out & 0.961 & 0.669 & 0.649 & 20 & 0 & 99 & 0.09 & 73\\ 
CU20L1B4Out & -4.563 & 0 & 0.036 & 0 & 20 & 0 & 91.26 & 0\\ 
CU30L1B6Out & -3.316 & 0 & 0.039 & 0 & 34 & 0 & 96.8 & 0\\ 
COMBINED & -3.76 & 0.107 & 0.203 & 54 & 69 & 673 & 61.75 & 582\\ 
 &  &  &  &  &  &  &  & \\ 
 \hline
DETR + EMD &  &  &  &  &  &  &  & \\ 
CU10L1B6In & -25.021 & 0.007 & 0.015 & 0 & 7 & 284 & 98.89 & 200\\ 
CU15L1B1In & 0.918 & 0.876 & 0.811 & 37 & 0 & 24 & 0.65 & 21\\ 
CU15L1B4In & -11.74 & 0.039 & 0.074 & 3 & 4 & 355 & 95.48 & 368\\ 
CU20L1B1Out & 0.97 & 0.712 & 0.696 & 21 & 0 & 66 & 0.27 & 50\\ 
CU20L1B4Out & -4.778 & 0 & 0.036 & 0 & 20 & 2 & 95.57 & 1\\ 
CU30L1B6Out & -3.372 & 0 & 0.039 & 0 & 34 & 0 & 98.43 & 0\\ 
COMBINED & -3.830 & 0.11 & 0.205 & 61 & 65 & 731 & 63.2 & 640\\ 
 &  &  &  &  &  &  &  & \\ 
  \hline

Transtrack &  &  &  &  &  &  &  & \\ 
CU10L1B6In & -24.859 & 0.003 & 0.01 & 0 & 26 & 158 & 97.71 & 151\\ 
CU15L1B1In & 0.833 & 0.519 & 0.527 & 40 & 1 & 144 & 1.37 & 309\\ 
CU15L1B4In & -11.842 & 0.011 & 0.026 & 0 & 24 & 390 & 93.92 & 421\\ 
CU20L1B1Out & 0.678 & 0.314 & 0.353 & 20 & 0 & 352 & 3.41 & 697\\ 
CU20L1B4Out & -3.869 & 0.016 & 0.029 & 0 & 3 & 1312 & 82.08 & 589\\ 
CU30L1B6Out & -2.976 & 0.012 & 0.027 & 0 & 30 & 854 & 89.05 & 379\\ 
COMBINED & -3.606 & 0.066 & 0.132 & 60 & 84 & 3210 & 59.76 & 2546\\ 
 &  &  &  &  &  &  &  & \\ 
  \hline

Trackformer &  &  &  &  &  &  &  & \\ 
CU10L1B6In & -11.199 & 0.001 & 0.008 & 0 & 30 & 52 & 43.93 & 65\\ 
CU15L1B1In & 0.918 & 0.829 & 0.733 & 36 & 1 & 27 & 0.66 & 23\\ 
CU15L1B4In & -6.254 & 0.011 & 0.024 & 1 & 35 & 311 & 49.23 & 469\\ 
CU20L1B1Out & 0.821 & 0.522 & 0.522 & 20 & 0 & 65 & 2.34 & 80\\ 
CU20L1B4Out & -0.42 & 0.023 & 0.019 & 0 & 18 & 600 & 9.55 & 623\\ 
CU30L1B6Out & -0.646 & 0.004 & 0.017 & 0 & 34 & 93 & 18.97 & 91\\ 
COMBINED & -1.027 & 0.187 & 0.262 & 57 & 118 & 1148 & 20.87 & 1351\\ 
 &  &  &  &  &  &  &  & \\ 
\hline

\bottomrule
\end{tabular}
}}
\end{center}

\caption{
Models trained only on source domain data
}
\label{comparisontable1}

\end{table}

\begin{table}[h]
  \begin{center}
    {\scriptsize{}{
    \setlength\tabcolsep{2pt}
\begin{tabular}{lrrrrrrrr}
\toprule
Sequence & MOTA & IDF1 & HOTA & MT & ML & Frag & FP(/frame) & ID Sw. \\
\midrule
DA FRCNN + SORT &  &  &  &  &  &  &  & \\ 
CU10L1B6In & -3.156 & 0.042 & 0.057 & 0 & 25 & 95 & 12.92 & 60\\ 
CU15L1B1In & 0.565 & 0.565 & 0.466 & 10 & 3 & 282 & 2.34 & 93\\ 
CU15L1B4In & -0.128 & 0.391 & 0.308 & 11 & 9 & 166 & 5.7 & 87\\ 
CU20L1B1Out & 0.84 & 0.513 & 0.443 & 15 & 0 & 212 & 0.02 & 105\\ 
CU20L1B4Out & 0.38 & 0.146 & 0.139 & 2 & 1 & 594 & 1.8 & 393\\ 
CU30L1B6Out & 0.034 & 0.082 & 0.068 & 0 & 19 & 525 & 4.41 & 348\\ 
COMBINED & 0.216 & 0.303 & 0.288 & 38 & 57 & 1874 & 4.53 & 1086\\ 
 &  &  &  &  &  &  &  & \\ 
 \hline

DA FRCNN + EMD &  &  &  &  &  &  &  & \\ 
]CU10L1B6In & -3.347 & 0.059 & 0.074 & 0 & 16 & 218 & 14.07 & 91\\ 
CU15L1B1In & 0.542 & 0.618 & 0.508 & 15 & 1 & 494 & 2.93 & 31\\ 
CU15L1B4In & -0.163 & 0.409 & 0.32 & 18 & 4 & 316 & 6.73 & 111\\ 
CU20L1B1Out & 0.87 & 0.569 & 0.489 & 15 & 0 & 291 & 0.04 & 109\\ 
CU20L1B4Out & 0.456 & 0.187 & 0.179 & 3 & 0 & 1040 & 2.35 & 407\\ 
CU30L1B6Out & 0.052 & 0.099 & 0.09 & 1 & 10 & 1043 & 6.06 & 474\\ 
COMBINED & 0.225 & 0.333 & 0.316 & 52 & 31 & 3402 & 5.8 & 1223\\ 
 &  &  &  &  &  &  &  & \\ 
\hline

\bottomrule
\end{tabular}
}}
\end{center}

\caption{
DA-FRCNN trained on source domain images, labels and target domain images
}
\label{comparisontable2}

\end{table}

\begin{table}[h]
  \begin{center}
    {\scriptsize{}{
    \setlength\tabcolsep{2pt}
\begin{tabular}{lrrrrrrrr}
\toprule
Sequence & MOTA & IDF1 & HOTA & MT & ML & Frag & FP(/frame) & ID Sw. \\
\midrule
Adapt DETR + SORT &  &  &  &  &  &  &  & \\ 
CU10L1B6In & -1.33 & 0.139 & 0.132 & 0 & 10 & 151 & 6.57 & 83\\ 
CU15L1B1In & 0.854 & 0.475 & 0.473 & 31 & 1 & 188 & 0.69 & 205\\ 
CU15L1B4In & 0.431 & 0.564 & 0.477 & 42 & 2 & 79 & 3.36 & 51\\ 
CU20L1B1Out & 0.657 & 0.2 & 0.258 & 19 & 1 & 352 & 4.24 & 541\\ 
CU20L1B4Out & 0.482 & 0.136 & 0.14 & 4 & 2 & 664 & 0.32 & 537\\ 
CU30L1B6Out & 0.15 & 0.08 & 0.067 & 0 & 24 & 417 & 0.32 & 306\\ 
COMBINED & 0.411 & 0.254 & 0.284 & 96 & 40 & 1851 & 2.7 & 1723\\ 
 &  &  &  &  &  &  &  & \\ 
 \hline
Adapt DETR + EMD &  &  &  &  &  &  &  & \\ 
CU10L1B6In & -1.328 & 0.207 & 0.176 & 4 & 4 & 289 & 7.27 & 72\\ 
CU15L1B1In & 0.809 & 0.466 & 0.482 & 36 & 1 & 24 & 1.87 & 202\\ 
CU15L1B4In & 0.403 & 0.576 & 0.487 & 42 & 2 & 70 & 3.88 & 46\\ 
CU20L1B1Out & 0.517 & 0.21 & 0.267 & 20 & 1 & 97 & 7.81 & 560\\ 
CU20L1B4Out & 0.579 & 0.184 & 0.189 & 8 & 1 & 980 & 1.15 & 638\\ 
CU30L1B6Out & 0.231 & 0.117 & 0.103 & 2 & 12 & 1053 & 1.05 & 567\\ 
COMBINED & 0.413 & 0.275 & 0.305 & 112 & 21 & 2513 & 4.05 & 2085\\ 
 &  &  &  &  &  &  &  & \\ 
 \hline
DA-Tracker &  &  &  &  &  &  &  & \\ 
CU10L1B6In & -0.145 & 0.181 & 0.141 & 0 & 10 & 135 & 1.74 & 134\\ 
CU15L1B1In & 0.777 & 0.812 & 0.695 & 20 & 5 & 23 & 0.27 & 5\\ 
CU15L1B4In & 0.197 & 0.493 & 0.413 & 11 & 13 & 23 & 3.67 & 14\\ 
CU20L1B1Out & 0.627 & 0.656 & 0.564 & 7 & 5 & 13 & 0.01 & 10\\ 
CU20L1B4Out & 0.594 & 0.42 & 0.31 & 4 & 0 & 151 & 0.85 & 99\\ 
CU30L1B6Out & 0.321 & 0.213 & 0.141 & 0 & 9 & 410 & 0.94 & 233\\ 
COMBINED & 0.493 & 0.494 & 0.433 & 42 & 42 & 755 & 1.26 & 495\\ 
 &  &  &  &  &  &  &  & \\ 
\hline

\bottomrule
\end{tabular}
}}
\end{center}

\caption{
Our models trained on source domain images, labels and target domain images
}
\label{comparisontable3}

\end{table}

\begin{table}[t]
  \begin{center}
    {\scriptsize{}{
    \setlength\tabcolsep{2pt}
\begin{tabular}{lrrrrrrrr}
\toprule
Sequence & MOTA & IDF1 & HOTA & MT & ML & Frag & FP(/frame) & ID Sw. \\
\midrule
Transtrack &  &  &  &  &  &  &  & \\ 
CU10L1B6In & 0.865 & 0.7 & 0.577 & 30 & 1 & 32 & 0.37 & 49\\ 
CU15L1B1In & 0.95 & 0.718 & 0.696 & 40 & 0 & 52 & 0.33 & 108\\ 
CU15L1B4In & 0.948 & 0.786 & 0.732 & 56 & 0 & 46 & 0.2 & 37\\ 
CU20L1B1Out & 0.938 & 0.628 & 0.64 & 21 & 0 & 101 & 0.53 & 182\\ 
CU20L1B4Out & 0.94 & 0.509 & 0.545 & 20 & 0 & 167 & 0.34 & 154\\ 
CU30L1B6Out & 0.859 & 0.425 & 0.405 & 33 & 0 & 421 & 1.48 & 293\\ 
COMBINED & 0.918 & 0.584 & 0.584 & 200 & 1 & 819 & 0.49 & 823\\ 
 &  &  &  &  &  &  &  & \\ 
 \hline
Trackformer &  &  &  &  &  &  &  & \\ 
CU10L1B6In & 0.852 & 0.849 & 0.643 & 28 & 1 & 66 & 0.24 & 19\\ 
CU15L1B1In & 0.941 & 0.872 & 0.794 & 36 & 1 & 19 & 0.37 & 10\\ 
CU15L1B4In & 0.875 & 0.815 & 0.725 & 50 & 2 & 24 & 0.47 & 8\\ 
CU20L1B1Out & 0.967 & 0.752 & 0.689 & 20 & 0 & 22 & 0.05 & 27\\ 
CU20L1B4Out & 0.943 & 0.534 & 0.522 & 18 & 0 & 91 & 0.13 & 110\\ 
CU30L1B6Out & 0.841 & 0.394 & 0.375 & 28 & 0 & 393 & 0.74 & 289\\ 
COMBINED & 0.912 & 0.646 & 0.61 & 180 & 4 & 615 & 0.31 & 463 \\
\hline

\bottomrule
\end{tabular}
}}
\end{center}

\caption{
Joint-detection-tracking models trained target domain images and labels
}
\label{comparisontable4}

\end{table}

\begin{figure}[tb]
\begin{center}
\includegraphics[width=0.5\textwidth]{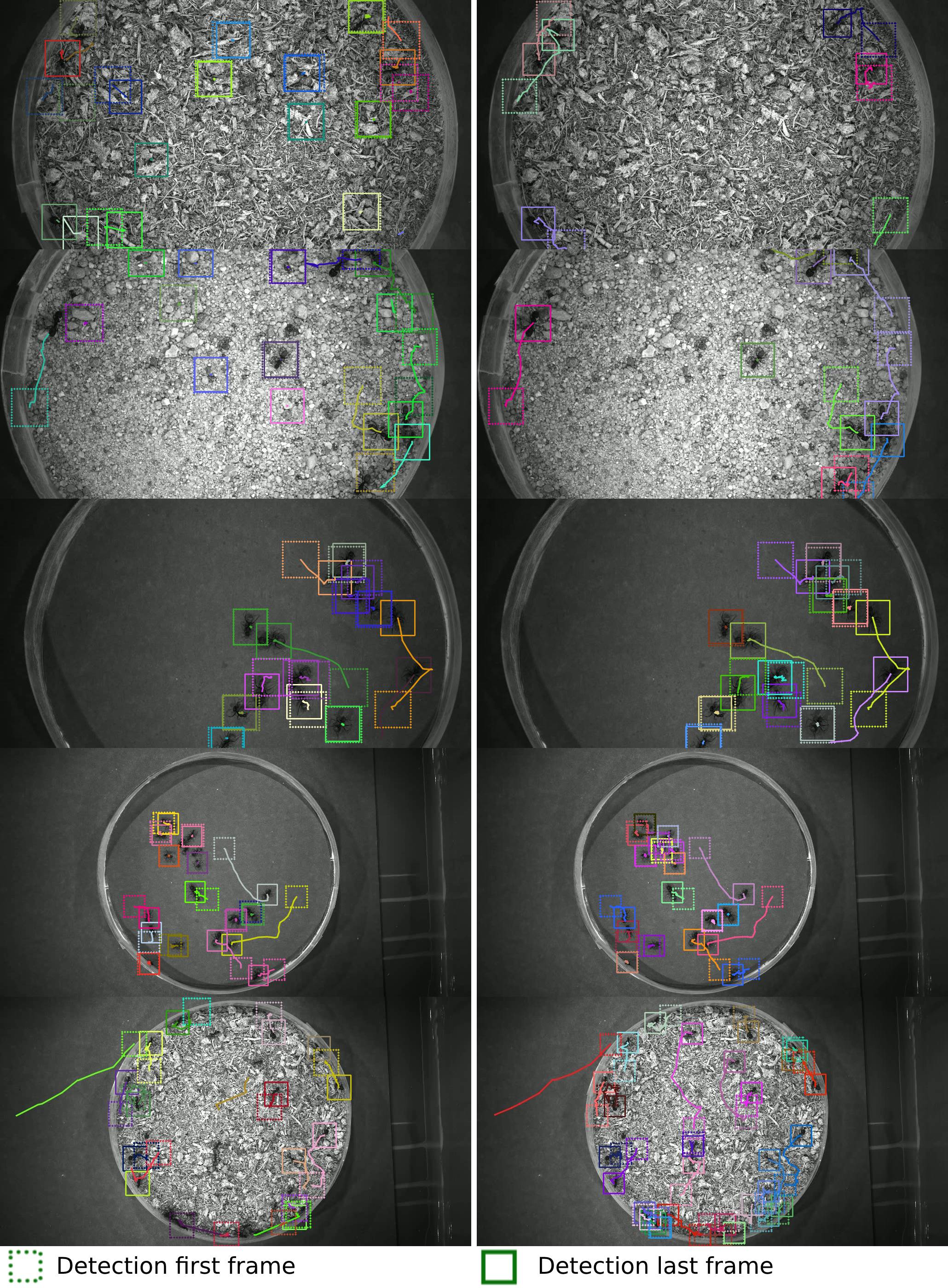}
\end{center}
   \caption{Tracking predictions for 20 frames from our DA-Tracker (left) and Ground truth (right)}
\label{fig:aditionalimages}
\end{figure}

\section*{Results}

The accuracy of our models at the sequence level is displayed in Tab. \ref{comparisontable1},\ref{comparisontable2},\ref{comparisontable3} and \ref{comparisontable4}. These results are an expansion of what was previously presented in our paper. Fig. \ref{fig:aditionalimages} provides visualizations of DA-Tracker prediction and ground truth for other test videos in our dataset.

\section*{Video resources}
The supplementary material includes a video that demonstrates a comparison of our results to those of the Trackformer baseline, as well as the results of Trackformer trained on the target domain and samples from the dataset. Video is available at \textcolor{magenta}{\emph{https://bit.ly/da-tracker}}.

\section*{Dataset}
Table \ref{datsetinfotable1} and \ref{datsetinfotable2} provides a detailed view about the proposed dataset.

\begin{table}[h]
  \begin{center}
    {\scriptsize{}{
    \setlength\tabcolsep{2pt}
\begin{tabular}{lrrrrrrrr}
\toprule
Sequence & Duration & Frames & Detection & Tracks & Avg. track length \\
\midrule
Train split & & & & \\
OU10B1L1Out & 42 & 500 & 5000 & 10 & 500.0 \\
OU10B1L2Out & 42 & 500 & 5000 & 10 & 500.0 \\
OU10B1L3In & 42 & 500 & 3223 & 16 & 201.4 \\
OU10B2L1Out & 60 & 720 & 7200 & 10 & 720.0 \\
OU10B2L2Out & 60 & 720 & 7200 & 10 & 720.0 \\
OU10B2L3In & 60 & 720 & 7110 & 11 & 646.4 \\
OU10B2L3Out & 60 & 720 & 6480 & 9 & 720.0 \\
OU10B3L2In & 60 & 720 & 7187 & 10 & 718.7 \\
OU10B3L2Out & 60 & 720 & 7200 & 10 & 720.0 \\
OU10B3L3In & 60 & 720 & 7110 & 10 & 711.0 \\
OU10B3L3Out & 60 & 720 & 7200 & 10 & 720.0 \\
OU50B1L1In & 42 & 500 & 14972 & 69 & 217.0 \\
OU50B1L2Out & 42 & 500 & 22979 & 47 & 488.9 \\
OU50B1L3In & 42 & 500 & 17956 & 92 & 195.2 \\
OU50B2L1In & 60 & 720 & 32638 & 55 & 593.4 \\
OU50B2L2In & 42 & 500 & 23859 & 52 & 458.8 \\
OU50B2L3In & 60 & 720 & 33383 & 53 & 629.9 \\
OU50B3L1In & 42 & 500 & 22572 & 50 & 451.4 \\
OU50B3L2In & 42 & 500 & 22992 & 52 & 442.2 \\
OU50B3L3Out & 42 & 500 & 24500 & 49 & 500.0 \\
\hline
Validation split & & & & \\
OU10B1L1In & 42 & 500 & 3754 & 26 & 144.4 \\
OU10B2L2In & 60 & 720 & 6210 & 11 & 564.5 \\
OU10B3L1Out & 60 & 720 & 7200 & 10 & 720.0 \\
OU50B1L2In & 42 & 500 & 17648 & 72 & 245.1 \\
OU50B1L3Out & 42 & 500 & 24500 & 49 & 500.0 \\
OU50B3L3In & 42 & 500 & 23105 & 49 & 471.5 \\
\hline
Test split & & & & \\
OU10B1L2In & 42 & 500 & 1952 & 13 & 150.2 \\
OU10B1L3Out & 42 & 500 & 5000 & 10 & 500.0 \\
OU10B2L1In & 60 & 720 & 5097 & 17 & 299.8 \\
OU10B3L1In & 60 & 720 & 7120 & 11 & 647.3 \\
OU50B1L1Out & 42 & 500 & 23553 & 49 & 480.7 \\
OU50B2L2Out & 60 & 720 & 36000 & 50 & 720.0 \\
OU50B3L2Out & 42 & 500 & 24500 & 49 & 500.0 \\
\hline

\bottomrule
\end{tabular}
}}
\end{center}

\caption{
Source domain video sequences}
\label{datsetinfotable1}

\end{table}

\begin{table}[h]
  \begin{center}
    {\scriptsize{}{
    \setlength\tabcolsep{2pt}
\begin{tabular}{lrrrrrrrr}
\toprule
Sequence & Duration & Frames & Detection & Tracks & Avg. track length \\
\midrule
Train split & & & & \\
CU10L1B1In & 60 & 720 & 6671 & 20 & 333.6 \\
CU10L1B1Out & 60 & 720 & 7200 & 10 & 720.0 \\
CU10L1B4In & 60 & 720 & 4788 & 43 & 111.3 \\
CU10L1B4Out & 60 & 720 & 7200 & 10 & 720.0 \\
CU10L1B5In & 60 & 720 & 4049 & 28 & 144.6 \\
CU10L1B6Out & 60 & 720 & 7200 & 10 & 720.0 \\
CU25L1B1In & 60 & 720 & 14451 & 50 & 289.0 \\
CU25L1B1Out & 60 & 720 & 16757 & 25 & 670.3 \\
CU25L1B4In & 60 & 720 & 6998 & 69 & 101.4 \\
CU25L1B4Out & 60 & 720 & 17280 & 24 & 720.0 \\
CU50L1B6In & 42 & 500 & 17581 & 116 & 151.6 \\
CU50L1B6Out & 42 & 500 & 24977 & 50 & 499.5 \\
\hline
Validation split & & & & \\
CU10L1B5Out & 60 & 720 & 7200 & 10 & 720.0 \\
CU15L1B1Out & 60 & 720 & 10261 & 15 & 684.1 \\
CU15L1B4Out & 60 & 720 & 10800 & 15 & 720.0 \\
CU20L1B1In & 60 & 720 & 9009 & 53 & 170.0 \\
CU20L1B4In & 60 & 720 & 9128 & 63 & 144.9 \\
CU30L1B6In & 42 & 500 & 7309 & 59 & 123.9 \\
\hline
Test split & & & & \\
CU10L1B6In & 56 & 666 & 2822 & 31 & 91.0 \\
CU15L1B1In & 60 & 720 & 9752 & 41 & 237.9 \\
CU15L1B4In & 60 & 720 & 5610 & 56 & 100.2 \\
CU20L1B1Out & 60 & 720 & 13409 & 21 & 638.5 \\
CU20L1B4Out & 60 & 720 & 14400 & 20 & 720.0 \\
CU30L1B6Out & 42 & 500 & 14595 & 34 & 429.3 \\
\hline
\bottomrule
\end{tabular}
}}
\end{center}

\caption{Target domain video sequences}
\label{datsetinfotable2}

\end{table}